\documentclass[10pt,twocolumn,letterpaper]{article}

\usepackage{cvpr}              
\usepackage{graphicx}
\usepackage{amsmath}
\usepackage{amssymb}
\usepackage{booktabs}
\usepackage{multirow}
\usepackage{bbding}
\usepackage{array}
\usepackage[title]{appendix}
\usepackage[group-separator={,}, group-minimum-digits=4]{siunitx}
\usepackage{tablefootnote}
\usepackage[para,symbol]{footmisc}

\usepackage[pagebackref,breaklinks,colorlinks]{hyperref}

\usepackage[pagebackref,breaklinks,colorlinks]{hyperref}

\usepackage[capitalize]{cleveref}
\crefname{section}{Sec.}{Secs.}
\Crefname{section}{Section}{Sections}
\Crefname{table}{Table}{Tables}
\crefname{table}{Tab.}{Tabs.}

\def\cvprPaperID{3338} 
\def\confName{CVPR}
\def\confYear{2023}
\makeatletter
 \def\@fnsymbol#1{\ensuremath{%
    \ifcase#1
    \or 
      \dagger
    \or 
      \ddagger
    \or 
      \mathsection
    \or 
      \mathparagraph
    \else 
      \@ctrerr  
    \fi}}   
\makeatother 

\begin{document}

\title{Contrastive Grouping with Transformer for Referring Image Segmentation}

\author{
Jiajin Tang$^{1}$ \quad 
Ge Zheng$^{1}$ \quad
Cheng Shi$^{1}$ \quad
Sibei Yang$^{1,2}$\thanks{Sibei Yang is the corresponding author.} \vspace{2mm}\\
$^1$School of Information Science and Technology, ShanghaiTech University \\
$^2$Shanghai Engineering Research Center of Intelligent Vision and Imaging \\
 {\tt\small \{tangjj,zhengge,shicheng2022,yangsb\}@shanghaitech.edu.cn} \vspace{-0mm}
}

\maketitle

\begin{abstract}
Referring image segmentation aims to segment the target referent in an image conditioning on a natural language expression. Existing one-stage methods employ per-pixel classification frameworks, which attempt straightforwardly to align vision and language at the pixel level, thus failing to capture critical object-level information. In this paper, we propose a mask classification framework, Contrastive Grouping with Transformer network (CGFormer), which explicitly captures object-level information via token-based querying and grouping strategy. Specifically, CGFormer first introduces learnable query tokens to represent objects and then alternately queries linguistic features and groups visual features into the query tokens for object-aware cross-modal reasoning. In addition, CGFormer achieves cross-level interaction by jointly updating the query tokens and decoding masks in every two consecutive layers. Finally, CGFormer cooperates contrastive learning to the grouping strategy to identify the token and its mask corresponding to the referent. Experimental results demonstrate that CGFormer outperforms state-of-the-art methods in both segmentation and generalization settings consistently and significantly. Code is available at \url{https://github.com/Toneyaya/CGFormer}.
\end{abstract}

\begin{figure}[t]
\centering
\includegraphics[width=0.95\columnwidth]{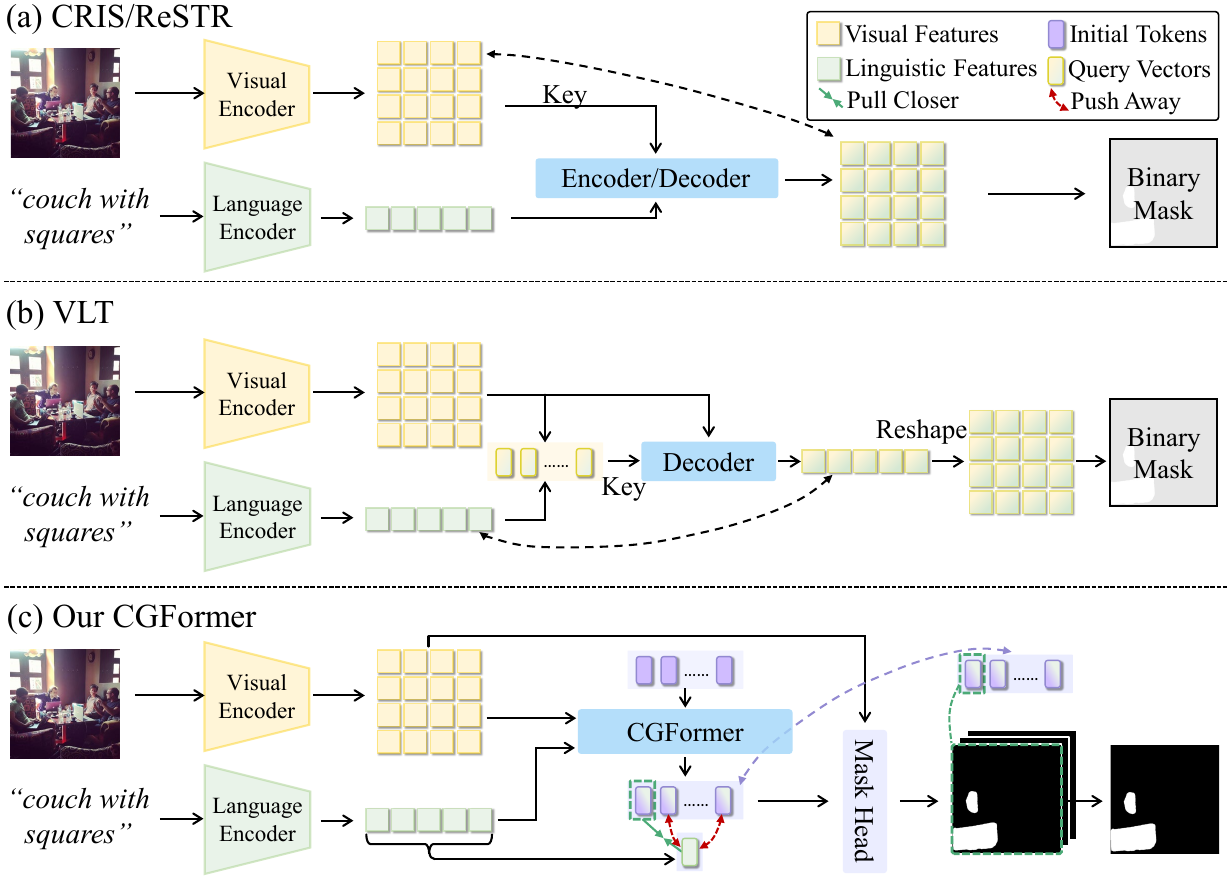}
\caption{Comparison of Transformer-based RIS methods and our CGFormer. (a) CRIS~\cite{wang2022cris} and ReSTR~\cite{kim2022restr} fuse relevant linguistic features into visual features, while (b) VLT~\cite{ding2021vision} generates query vectors to query visual features for segmentation. In contrast, (c) our CGFormer introduces learnable query tokens and explicitly groups visual features into tokens conditioning on language. Cooperating the grouping strategy with contrastive learning, it identifies the token and its mask corresponding to the referent.}
\label{fig:intro}
\vspace{-0.2cm}
\end{figure}

\section{Introduction}
\label{sec:intro}

Referring Image Segmentation (RIS) aims to segment the target referent in an image given a natural language expression~\cite{hu2016segmentation, yu2018mattnet, ding2021vision}. It attracts increasing attention in the research community and is expected to show its potential in real applications, such as human-robot interaction via natural language~\cite{wang2019reinforced} and image editing~\cite{chen2018language}. Compared to classical image segmentation that classifies pixels or masks into a closed set of fixed categories, RIS requires locating referent at pixel level according to the free-form natural languages with open-world vocabularies. It faces challenges of comprehensively understanding vision and language modalities and aligning them at pixel level. 

Existing works mainly follow the segmentation framework of \textit{per-pixel classification}~\cite{hu2016segmentation, long2015fully} integrated with multi-modal fusion to address the challenges. They introduce various fusion methods~\cite{li2018referring,liu2017recurrent,shi2018key,ye2019cross,hui2020linguistic,huang2020referring} to obtain vision-language feature maps and predict the segmentation results based on the feature maps.  
Recently, the improvement of vision-language fusion for RIS mainly lies in utilizing the Transformer~\cite{yang2022lavt,wang2022cris,kim2022restr,ding2021vision}. LAVT~\cite{yang2022lavt} integrates fusion module into the Transformer-based visual encoder. CRIS~\cite{wang2022cris} and ReSTR~\cite{kim2022restr} fuse linguistic features into each feature of the visual feature maps, as shown in Figure~\ref{fig:intro}\textcolor{red}{a}. In contrast, VLT~\cite{ding2021vision} integrates the relevant visual features into language-conditional query vectors via transformer decoder, as shown in Figure~\ref{fig:intro}\textcolor{red}{b}. 

Although these methods have improved the segmentation accuracy, they still face several intrinsic limitations. First, the works~\cite{yang2022lavt,wang2022cris,kim2022restr} based on pixel-level fusion only model the pixel-level dependencies for each visual feature, which fails to capture the crucial object/region-level information. Therefore, they cannot accurately ground expressions that require efficient cross-modal reasoning on objects. Second, although VLT~\cite{ding2021vision}'s query vectors contain object-level information after querying, it directly weights and reshapes different tokens into one multi-modal feature map for decoding the final segmentation mask. Therefore, it loses the image's crucial spatial priors (relative spatial arrangement among pixels) in the reshaping process. More importantly, it does not model the inherent differences between query vectors, resulting in that even though different query vectors comprehend expressions in their own way, they still focus on similar regions but fail to focus on different regions and model their relations.

In this paper, we aim to propose a simple and effective framework to address these limitations. Instead of using \textit{per-pixel classification} framework, we adopt an end-to-end \textit{mask classification} framework~\cite{he2017mask,carion2020end} (see Figure~\ref{fig:intro}\textcolor{red}{c}) to explicitly capture object-level information and decode segmentation masks for both the referent and other disturbing objects/stuffs. Therefore, we can simplify RIS task by finding the corresponding mask for the expression. Note that our framework differs from two-stage RIS methods~\cite{wu2020phrasecut,yu2018mattnet} which require explicitly detecting the objects first and then predicting the mask in the detected bounding boxes.

Specifically, we propose a Contrastive Grouping with Transformer (\textcolor{black}{CGFormer}) network consisting of the \textit{Group Transformer} and \textit{Consecutive Decoder} modules. The \textit{Group Transformer} aims to capture object-level information and achieve object-aware cross-modal reasoning. The success of applying query tokens in object detection~\cite{carion2020end,zhu2020deformable} and instance segmentation~\cite{he2017mask,cheng2021mask2former,wu2022seqformer} could be a potential solution. However, it is non-trivial to apply them to RIS. Without the annotation supervision of other mentioned objects other than the referent, it is hard to make tokens pay attention to different objects and distinguish the token corresponding to the referent from other tokens. Therefore, although we also specify query tokens as object-level information representations, \textit{we explicitly group the visual feature map's visual features into query tokens to ensure that different tokens focus on different visual regions without overlaps.} 
Besides, we can further \textit{cooperate contrastive learning with the grouping strategy} to make the referent token attend to the referent-relevant information while forcing other tokens to focus on different objects and background regions, as shown in Figure~\ref{fig:intro}\textcolor{red}{c}. In addition, we alternately query the linguistic features and group the visual features into the query tokens for cross-modal reasoning. 

Furthermore, integrating and utilizing multi-level feature maps are crucial for accurate segmentation. Previous works\cite{liu2017recurrent, margffoy2018dynamic,ye2019cross} fuse visual and linguistic features at multiple levels in parallel and later integrate them via ConvLSTM~\cite{shi2015convolutional} or FPNs~\cite{lin2017feature}. However, their fusion modules are solely responsible for cross-modal alignment at each level, which fails to perform joint reasoning for multiple levels. Therefore, we propose a \textit{Consecutive Decoder} that jointly updates query tokens and decodes masks in every two consecutive layers to achieve cross-level reasoning. 

To evaluate the effectiveness of CGFormer, we conduct experiments on three standard benchmarks, \ie, RefCOCO series datasets~\cite{yu2016modeling,mao2016generation,nagaraja2016modeling}. In addition, unlike semantic segmentation, RIS is not limited by the close-set classification but to open-vocabulary alignment. It is necessary to evaluate the generalization ability of RIS models. Therefore, we introduce new subsets of training sets on the three datasets to ensure the categories of referents in the test set are not seen in the training stage, inspired by the zero-shot visual grounding~\cite{sadhu2019zero} and open-set object detection~\cite{zareian2021open}. 

In summary, our main contributions are as follows,
\begin{itemize}
\setlength{\itemsep}{0pt}
\setlength{\parsep}{0pt}
\setlength{\parskip}{0pt}
	\item We propose a Group Transformer cooperated with contrastive learning to achieve object-aware cross-modal reasoning by explicitly grouping visual features into different regions and modeling their dependencies conditioning on linguistic features.
	\item We propose a Consecutive Decoder to achieve cross-level reasoning and segmentation by jointly performing the cross-modal inference and mask decoding in every two consecutive layers in the decoder.  
	\item We are the first to introduce an \textcolor{black}{end-to-end} mask classification framework, the Contrastive Grouping with Transformer (CGFormer), for referring image segmentation. Experimental results demonstrate that our CGFormer outperforms all state-of-the-art methods on all three benchmarks consistently.
	\item We introduce new splits on datasets for evaluating generalization for referring image segmentation models. CGFormer shows stronger generalizability compared to state-of-the-art methods thanks to object-aware cross-modal reasoning via contrastive learning. 
\end{itemize}

\begin{figure*}[th]
\centering
\includegraphics[width=0.95\textwidth]{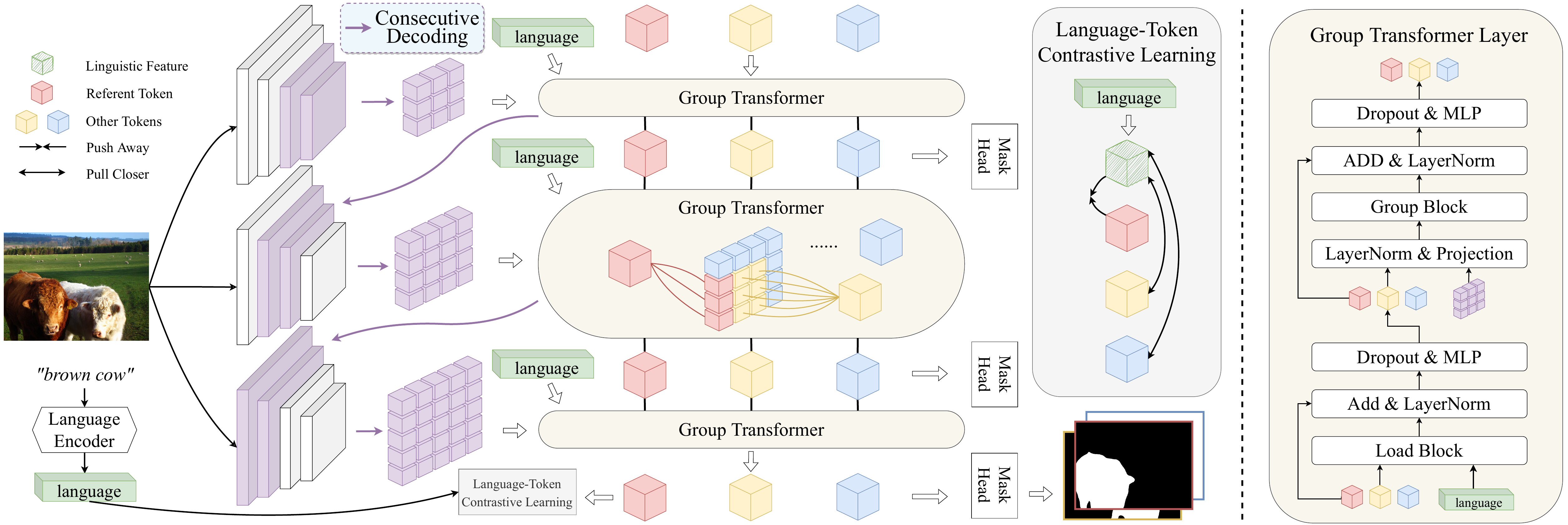}
\caption{Overall framework of the proposed CGFormer. We first extract visual and linguistic features and then feed them to Group Transformer to integrate multi-modal features to object-level tokens in a Consecutive Decoding way. Next, we distinguish the referent token from others via Contrastive Learning between language-token pairs and decode segmentation results for tokens via mask head.}
\label{fig:pipeline}
\vspace{-0.2cm}
\end{figure*}

\section{Related Work}
\label{sec:relat}
\noindent \textbf{Referring Image Segmentation} (RIS) aims to segment objects from images according to natural language expressions.
The pioneering work~\cite{hu2016segmentation} uses the concatenation operation to fuse the linguistic and visual features. 
Some following works~\cite{hu2016segmentation,li2018referring,chen2019referring,luo2020multi,hu2020bi,jing2021locate,jiao2021two} extract textual features for the expressions at the sentence level, while other works~\cite{liu2017recurrent,margffoy2018dynamic,chen2019see,feng2021encoder} employ word vectors as textual representations. Considering that natural language naturally contains structured information~\cite{yang2020propagating,shi2022spatial} that can be exploited to align with visual constituents, some methods explicitly decompose expressions into different components~\cite{wu2020phrasecut,hui2020linguistic,yang2021bottom,yang2020graph} or apply soft component division via the attention mechanisms~\cite{shi2018key,yu2018mattnet,ye2019cross,huang2020referring,fu2019dual,ding2021vision, yang2019dynamic}. The composed components are then aligned with visual constituents via the well-designed module networks~\cite{wu2020phrasecut,yu2018mattnet,yang2021bottom,yang2019cross,yang2020relationship} or attention mechanism~\cite{shi2018key,ye2019cross,lin2021structured} and interact with each other
via graph convolution networks~\cite{hui2020linguistic,huang2020referring} or the transformer~\cite{ding2021vision}.

Recently, the research interest has shifted toward developing a better framework for vision-language fusion. 
LAVT~\cite{yang2022lavt} adopts Swin Transformer~\cite{liu2021swin} as the visual encoder and integrates vision-language fusion modules at the last four encoding layers in the visual encoder. \textcolor{black}{Alternatively, ReSTR~\cite{kim2022restr} and CRIS~\cite{wang2022cris}, which first encode vision and language with a dual encoder and then fuse visual and linguistic features by resorting to a multi-modal transformer encoder or cross-modal decoder.} Unlike existing one-stage RIS studies that are based on per-pixel classification framework, we convert the pixel-level alignment to the mask-level by selecting the mask corresponding to the expression. 

\noindent \textbf{Semantic and Instance Segmentation}. Semantic segmentation aims to segment regions according to visual semantics by labeling every pixels~\cite{long2015fully}. Mainstream methods adopt the segmentation framework of per-pixel classification. Specifically, FCNs~\cite{long2015fully} adopt a stack of convolutional blocks to classify pixels. Further, ASPP~\cite{chen2017deeplab,chen2017rethinking} and GCN~\cite{peng2017large} are applied to improve FCNs with larger receptive fields. Transformer-based models~\cite{strudel2021segmenter,zheng2021rethinking} further capture long-range dependencies. Unlike semantic segmentation, instance segmentation requires predicting both the masks and categories at the instance level~\cite{li2017fully}. To achieve this goal, mainstream methods adopt the segmentation framework of mask classification.
Specifically, Mask R-CNN~\cite{he2017mask} employs a two-stage framework, which first generates a set of proposals and then predicts the masks and categories for proposals. 
Moreover, DETR~\cite{carion2020end} adopts an end-to-end segmentation framework that uses a large number of learnable query tokens to represent instances and predicts the mask and category of the instance based on each corresponding token. Recently, MaskFormer~\cite{cheng2021per} expands the DETR and can be applied to both semantic and instance segmentation. 

To address RIS, we further exploit the advantages of per-pixel and mask classification by using the hard assignment to ensure that each pixel can only be grouped into one query token and avoid the overlap between the tokens' masks.

\noindent \textbf{Hard Assignment} is a reparameterization method to solve the problem of non-differentiable argmax operation. In recent works, Gumbel-Softmax~\cite{jang2016categorical,maddison2016concrete}, a hard assignment method, has been applied to semantic segmentation to group pixels with similar semantics~\cite{xu2022groupvit,yu2022k}. For example, GroupViT~\cite{xu2022groupvit} initializes several sets of queries representing multiple level semantics and uses Gumbel-Softmax to group pixels from low level to high level.
Similar to GroupViT, K-means Mask Transformer~\cite{yu2022k} generates several queries as clustering centers and clusters pixels with similar semantics via Gumbel-Softmax.

However, bottom-up clustering based on semantics is not applicable in RIS because RIS requires distinguishing the target region and disturbing regions with similar semantics. Therefore, instead of utilizing Gumbel-Softmax to group pixels from bottom to up, we use it to compare between our specialized tokens, which can avoid the target region and the disturbing regions being grouped together.

\section{Method}
\label{sec:method}
The framework of our proposed CGFormer is shown in Figure \ref{fig:pipeline}. First, we adopt the visual encoder and language encoder to encode images and referring expressions (see Section~\ref{sec:encoder}). Second, we achieve object-aware cross-modal reasoning via the proposed Group Transformer (see Section~\ref{sec:cg}). Next, we implement the cross-level reasoning and segmentation via the proposed Consecutive Decoder (see Section~\ref{sec:cd}). Finally, we apply contrastive learning to distinguish the referent token from other tokens and obtain the mask corresponding to the referent token as the segmentation result (see Section~\ref{sec:loss}).

\subsection{Visual Encoder and Language Encoder}
\label{sec:encoder}

\noindent\textbf{Visual Encoder.} Following the previous work~\cite{yang2022lavt}, we employ Swin Transformer~\cite{liu2021swin} as the visual encoder for fair comparison. For an input image $I\in \mathbb{R} ^{H\times W\times 3}$ with the size of $H \times W$, we extract its visual feature maps at four stage $i\in\{1,2,3,4\}$, which we denote it as $\boldsymbol{V}=\{V_i\}_{i=1}^4, V_i\in \mathbb{R}^{H_i\times W_i\times C_i^v}$. Here, each stage corresponds to an encoding block of Swin Transformer, and $H_i$, ${W_i}$ and $C_i^v$ denote the height, width and channel dimension of $V_i$.

\noindent\textbf{Language Encoder.}
We adopt BERT~\cite{devlin2018bert} as the language encoder following the previous work~\cite{yang2022lavt}. Given an expression contains $L$ words, we extract its linguistic feature and denote it as $\boldsymbol{e}\in\mathbb{R}^{C^l}$, where $C^l$ is the channel dimension. In addition, we obtain the word representations by removing the last pooling layer, which is denoted as $F \in \mathbb{R}^{L \times C^l}$.

\subsection{Group Transformer}
\label{sec:cg}
We propose the Group Transformer to achieve object-aware cross-modal inference. Group Transformer uses query tokens to represent object-level information and updates query tokens by alternately querying the linguistic features and grouping visual features. Specifically, we first initialize a set of learnable tokens representing the different objects/regions (see Section \ref{sec:init}). Then we query the linguistic and visual features for tokens via our proposed novel Group Transformer layer (see Section \ref{sec:layer}). After alternative reasoning, tokens capture the rich object characteristics relevant to the referring expressions. Note that we use multiple Group Transformer layers in different stages of the decoder, which will be introduced in Section \ref{sec:cdlayer}. 

\vspace{-0.3cm}
\subsubsection{Definition of Query Token}
\vspace{-0.1cm}
\label{sec:init}
Inspired by the apply of query tokens~\cite{carion2020end,wu2022seqformer,cheng2021per} in object detection and semantic segmentation, we randomly initialize $N$ learnable tokens $T \in \mathbb{R}^{N \times C^t}$ 
where $C^t$ is the channel dimension of tokens, to represent referent and other disturbing objects/stuffs. 

Next, we feed these tokens $T$ to Group Transformer layers to capture the object-level information conditioning on the expression and update the features for tokens. For simplicity of demonstration, we use $T_{i-1} \in \mathbb{R}^{N \times C^t}$ to represent the features of tokens output from the $(i-1)$-th layer and input them to the $i$-th layer of the Group Transformer.

\vspace{-0.2cm}
\subsubsection{Group Transformer Layer}
\vspace{-0.1cm}
\label{sec:layer}
The right part of Figure \ref{fig:pipeline} illustrates the architecture of a single Group Transformer layer. The Load block and Group block are two core blocks to achieve object-aware cross-modal reasoning. In addition, we follow standard transformer~\cite{vaswani2017attention} to employ $\mathrm{LayerNorm}$ for feature normalization and $\mathrm{MLP}$ with activation function for nonlinear mapping.
Specifically, the Load block preloads the linguistic information each token should focus on at the current layer. The Group block, the critical component of Group Transformer, performs the cross-modal interaction and groups the visual features into query tokens to ensure that different tokens focus on different visual regions without overlap.

\noindent \textbf{Load Block} is expected to preload what linguistic information the query tokens should focus on at the current layer. The load layer is implemented by a classical cross-attention block~\cite{vaswani2017attention}, accepting the input tokens $T_{i-1} \in \mathbb{R}^{N \times C^t}$ as the query and the word vectors $F \in \mathbb{R}^{L \times C^l}$ extracted by the text encoder (see Section~\ref{sec:encoder}) as the key and value.
Concretely, the computation of the Load block is as follows: 
\begin{equation}\label{eq:1}
    \begin{aligned}
T^{q}_i &= T_{i-1}W_{q} , 
F^{k} = FW_{k} , 
F^{v} = FW_{v} ,  \\
T^l_i &=  (\mathrm{softmax}(\frac{T^{q}_{i}(F^{k})^\top}{\sqrt{C^l}})F^{v})W_{c}, 
    \end{aligned}
\end{equation}
where $W_q, W_c \in\mathbb{R}^{C^t\times C^t}$ and $W_k, W_v\in\mathbb{R}^{C^l\times C^t}$ are learnable projection matrices. And $T^{q}_i, F^{k}$ and $F^{v}$ are query, key, and value in the cross attention, respectively. We end up with linguistic-enhanced representations for tokens, $T^l_i\in\mathbb{R}^{N\times C^t}$, and feed them into the Group block to query and group the relevant visual features for tokens.

\noindent \textbf{Group Block} interacts between vision and language and groups visual features from the feature map into linguistic-enhanced query tokens $T^l_i$. We denote the feature map as $D_{i}\in \mathbb{R}^{H_i\times W_i\times C_i^v}$, which is fused from the feature maps in two consecutive layers in Consecutive Decoder (refer to Section~\ref{sec:cd} for details). Firstly, we project $T^l_i$ and $D_{i}$ into a common feature space:
\begin{equation}\label{eq:2}
    \begin{aligned}
T^{\prime}_i = T^{l}_iW_{t}, 
D^{\prime}_{i} = \mathrm{flatten}(D_{i})W_{d}, 
    \end{aligned}
\end{equation}
where $W_{t}\in\mathbb{R}^{C^t\times C^t}$ and $W_{d}\in\mathbb{R}^{C^v_i\times C^t}$ are learnable projection matrices, and $\mathrm{flatten}$ operation flattens the feature map $D_{i}$ into the visual feature with $H_iW_i$ vectors. Then, we calculate the similarities $S_{pixel}\in \mathbb{R}^{N\times H_iW_i}$ between every pairwise features of tokens $T_i^\prime$ and features $D_i^\prime$:
\begin{equation}\label{eq:3}
    \begin{aligned}
        S_{pixel} = \mathrm{norm}(T_i^\prime) \mathrm{norm}(D^{\prime}_{i})^\top,
    \end{aligned}
\end{equation}
where $\mathrm{norm}$ means L2 normalization for vectors. 

Next, based on the similarities $S_{pixel}$, we group the features in $D_{i}^\prime$ and correspond the groups to tokens $T_i^\prime$. However, the grouping operation with straightforward hard assignment is non-differentiable. Therefore, we adopt a learnable Gumbel-softmax~\cite{jang2016categorical,maddison2016concrete} to hard assign the features in $D_{i}^\prime$ to the tokens $T_i^\prime$ and generate the mask $S_{mask}\in \mathbb{R}^{N\times H_iW_i}$ of the grouping. The computation is as follows:
\begin{equation}\label{eq:4}
    \begin{aligned}
    &S_{gumbel} = \mathrm{softmax}((S_{pixel}+G)/\mathop{\tau}\limits_{\ }),\\
    &S_{onehot} = \mathrm{onehot}(\mathrm{argmax}_N(S_{gumbel})),\\
    &S_{mask} = (S_{onehot})^\top-\mathrm{sg}(S_{gumbel})+S_{gumbel},\\
    \end{aligned}
\end{equation}
where $G\in \mathbb{R}^{N\times H_iW_i}$ samples from the $\mathrm{Gumbel}(0,1)$ distribution, $\tau$ is the learnable significance coefficient to assist in finding a more suitable assign boundary, $\mathrm{sg}$ is the stop gradient operator. Here, $\mathrm{argmax}_{N}$ means selecting the corresponding token of $T_i^\prime$ with the highest similarity for each feature in $D_{i}^\prime$, and the $\mathrm{onehot}$ operation transforms the token indexes into $H_iW_i$ one-hot vectors $S_{onehot}\in \mathbb{R}^{H_iW_i\times N}$. The mask $S_{mask}\in \mathbb{R}^{N\times H_iW_i}$ indicates the grouping from the features $D_{i}^\prime$ to the tokens $T_i^\prime$.

Finally, we integrate the features $D_{i}^\prime$ to update tokens $T_i^\prime$ based on the mask $S_{mask}$, which is computed as follows:
\begin{equation}\label{eq:5}
        T_{i} = \mathrm{MLP}(S_{mask}D_i^\prime) +  T_i^\prime,
\end{equation}
where $\mathrm{MLP}$ is the multilayer perceptron. And $T_{i}$ are the updated features of tokens via the Group block, which capture the rich object/region characteristics relevant to the linguistic features.

\subsection{Consecutive Decoder}
\label{sec:cd}

\begin{figure}[t]
\centering
\includegraphics[width=0.95\columnwidth]{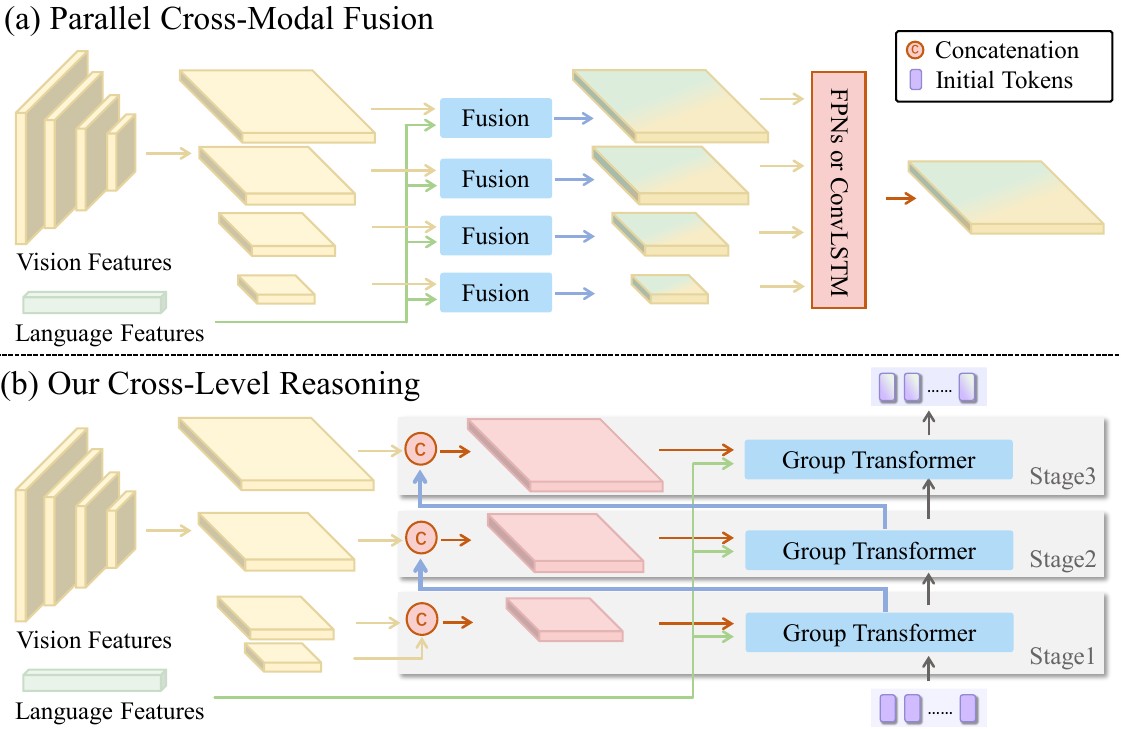}
\caption{Comparison of (a) parallel cross-modal fusion and (b) our cross-level reasoning via the Consecutive Decoder.} 
\label{fig:decoder}
\vspace{-0.1cm}
\end{figure}

We further perform cross-level reasoning via the proposed Consecutive Decoder. 
Figure~\ref{fig:decoder} shows the architecture of our Consecutive Decoder and its comparison to the parallel cross-modal fusion. 
Previous works~\cite{liu2017recurrent,margffoy2018dynamic,ye2019cross,chen2019see,hui2020linguistic,luo2020multi,hu2020bi,huang2020referring} model the vision-language interaction at multiple levels in parallel and late integrate multi-level results. The sole interaction at a single level fails to perform joint interaction across various levels. In contrast, the Consecutive Decoder achieves cross-level reasoning by jointly updating the query tokens in every two consecutive decoder layers, and the two-level cross-modal information will be consecutively propagated in multiple levels from bottom to up.

Specifically, the Consecutive Decoder contains three stages. At each decoding stage, it first fuses feature maps at two levels as the input of the Group Transformer layer to update tokens (see Section~\ref{sec:cdlayer}) and then decodes the corresponding mask of each token through the Mask Head (see Section~\ref{sec:mask}).

\vspace{-0.3cm}
\subsubsection{Consecutive Decoding}
\vspace{-0.1cm}
\label{sec:cdlayer}

We intersperse multi-scale and cross-modal reasoning at each decoding layer. 
Specifically, for the $i$-th decoder layer, we first adopt a convolutional module to fuse the visual feature map $V_i$ at the current layer $i$ and the multi-modal feature map $D_{i-1}$ output from the previous layer $i-1$ of the Consecutive Decoder to generate the feature map $D_{i}\in \mathbb{R}^{H_{i}\times W_{i}\times C_{i}^v}$. Then, we update the query tokens $T_{i-1}$ output from the previous Consecutive Decoder layer by querying them on the feature map $D_{i}$ via the Group Transformer layer. The calculation is as follows: 
\begin{equation}\label{eq:6}
    \begin{aligned}
D_{i} &= \mathrm{Conv}([V_{i}; \mathrm{Up}(D_{i-1})]), i \in \{2,3,4\}\\
T_{i} &= \mathrm{GroupTransformerLayer}(T_{i-1}, D_{i}),\\
    \end{aligned}
\end{equation}
where $\mathrm{Conv}$ is the convolution layer, $\mathrm{Up}$ refers to up-sampling $D_{i-1}$ to the scale of $V_i$, $[; ]$ denotes concatenation along the channel dimension. 

Particularly, for the first decoder layer ($i=1$), we skip the fusion and cross-modal interaction and let $D_1=V_1$ and $T_1=T$, where $T$ are the initialized tokens defined in Section~\ref{sec:init}.

\vspace{-0.3cm}
\subsubsection{Mask Head}
\vspace{-0.1cm}
\label{sec:mask}
For $i$-th decoder layer, our mask head takes updated tokens $T_i$ and visual feature map ${D_i}$ as inputs and output the segmentation probabilities $Z_i \in \mathbb{R}^{N \times H_i\times W_i}$ for tokens via dynamic convolutions~\cite{chen2020dynamic}. For $n$-th token with feature $T_i^{(n)}$, we first project it to convolution kernels $W_i^{(n)}$ and then predict the segmentation probabilities $Z_i^{(n)} \in \mathbb{R}^{H_i\times W_i}$ based on the kernels, which is computed as follows,
\begin{equation}\label{eq:9}
\begin{aligned}
    W_i^{(n)} &= \mathrm{MLP}(T_i^{(n)}), \\
    Z_{i}^{(n)} &= \mathrm{Sigmoid}(\mathrm{Conv}_{W_i^{(n)}}(D_i)),
\end{aligned}
\end{equation}
where the superscript $^{(n)}$ denotes the features, kernels, and predicted probabilities corresponding to the $n$-th token, and the $\mathrm{Conv}_{{W_i^{(n)}}}$ means the convolution layer with the convolution kernels ${W_i^{(n)}}$. 

\subsection{Contrastive Learning}
\label{sec:loss}

\begin{table*}[t]
   \centering
   \resizebox{1.85\columnwidth}{!}{
   \setlength{\tabcolsep}{3mm}{\begin{tabular}{l|l|c|c|c|c|c|c|c|c|c|c}
      \toprule[1pt]
      &\multirow{2}{*}{Method} &
      \multicolumn{3}{c|}{RefCOCO}  & \multicolumn{3}{c|}{RefCOCO+} & \multicolumn{3}{c|}{G-Ref}& \multicolumn{1}{c}{ReferIt} \\
      \cline{3-12}
      && val   & test A & test B & val & test A & test B & val-U & test-U & val-G & test \\
      \hline
      \multirow{7}{*}{mIoU}
      &DMN~\cite{margffoy2018dynamic}   & 49.78 & 54.83 & 45.13 & 38.88 & 44.22 & 32.29 & -      & -     & 36.76 & 52.81\\
      &MCN~\cite{luo2020multi}          & 62.44 & 64.20 & 59.71 & 50.62 & 54.99 & 44.69 & 49.22 & 49.40 & -  & -    \\
      &CGAN~\cite{luo2020cascade}       & 64.86 & 68.04 & 62.07 & 51.03 & 55.51 & 44.06 & 51.01 & 51.69 & 46.54 & - \\
      &LTS~\cite{jing2021locate}        & 65.43 & 67.76 & 63.08 & 54.21 & 58.32 & 48.02 & 54.40 & 54.25 & - & - \\
      &VLT~\cite{ding2021vision}        & 65.65 & 68.29 & 62.73 & 55.50 & 59.20 & 49.36 & 52.99 & 56.65 & 49.76 & - \\
      &CRIS~\cite{wang2022cris}         & 70.47 & 73.18 & 66.10 & 62.27 & 68.08 & 53.68 & 59.87 & 60.36 & -  & - \\
      \cline{2-12}
      \rule{0pt}{10pt} 
      &\textbf{Our CGFormer} & \textbf{76.93} & \textbf{78.70} & \textbf{73.32} & \textbf{68.56} & \textbf{73.76} & \textbf{61.72} & \textbf{67.57} & \textbf{67.83} & \textbf{65.79} & \textbf{66.42} \\
      \midrule[1pt]
      \multirow{10}{*}{oIoU}
      &RRN~\cite{li2018referring}       & 55.33 & 57.26 & 53.93 & 39.75 & 42.15 & 36.11 & -     & -     & 36.45 & 63.63 \\
      &MAttNet~\cite{yu2018mattnet}     & 56.51 & 62.37 & 51.70 & 46.67 & 52.39 & 40.08 & 47.64 & 48.61 & - & -     \\
      &CMSA~\cite{ye2019cross}          & 58.32 & 60.61 & 55.09 & 43.76 & 47.60 & 37.89 & -     & -     & 39.98 & 63.80 \\
      &CMPC~\cite{huang2020referring}   & 61.36 & 64.53 & 59.64 & 49.56 & 53.44 & 43.23 & -     & -     & 49.05 & 65.53 \\
      &LSCM~\cite{hui2020linguistic}    & 61.47 & 64.99 & 59.55 & 49.34 & 53.12 & 43.50 & -     & -     & 48.05 & 66.57  \\
      &CEFNet~\cite{feng2021encoder}    & 62.76 & 65.69 & 59.67 & 51.50 & 55.24 & 43.01 & 51.93 & - & -  & 66.70 \\ 
      &BUSNet~\cite{yang2021bottom}     & 63.27 & 66.41 & 61.39 & 51.76 & 56.87 & 44.13 & - & - & 50.56 & -  \\
      &ReSTR~\cite{kim2022restr}        & 67.22 & 69.30 & 64.45 & 55.78 & 60.44 & 48.27 & 54.48 & -     &  -   & -   \\
      &LAVT~\cite{yang2022lavt}         & 72.73 & 75.82 & 68.79 & 62.14 & 68.38 & 55.10 & 61.24 & 62.09 & 60.50 & - \\
      \cline{2-12}
      \rule{0pt}{10pt}
      &\textbf{Our CGFormer} & \textbf{74.75} & \textbf{77.30} & \textbf{70.64} & \textbf{64.54} & \textbf{71.00} & \textbf{57.14} & \textbf{64.68} & \textbf{65.09} & \textbf{62.51} & \textbf{73.36} \\
      \bottomrule[1pt]
   \end{tabular}}}
   \caption{Comparison with state-of-the-art models in referring image segmentation on RefCOCO, RefCOCO+, G-Ref, and ReferIt datasets.} 
   \label{tab:1}
\end{table*}
We use contrastive learning to distinguish the referent token from other tokens by maximizing the similarity between the referent token and the expression and minimizing the similarities between negative pairs. For simplicity of demonstration, we suppose the first token represents the referent and other tokens represent non-target objects/stuff. The contrastive loss between the tokens with features $T_4 \in \mathbb{R}^{N \times C^t}$ output from last decoder layer and the expression $\boldsymbol{e} \in \mathbb{R}^{C^l}$ is computed as follows, 
\begin{equation}\label{eq:10}
\mathcal{L}_{cl} = -\log(\frac{\text{exp}(s(T_4^{(1)}, \boldsymbol{e}))}{\sum_{n=2}^{N} \text{exp}(s(T_4^{(n)}, \boldsymbol{e}))}),
\end{equation}
where $s(\cdot,\cdot)$ is used to compute the similarity.

In addition, we combine the dice loss~\cite{li2019dice} and binary cross-entropy loss as the segmentation loss, $\mathcal{L}_{seg}$. And we use the segmentation loss on multiple levels to supervise the learning of masks $Z_i$. The total loss $\mathcal{L}$ is the sum of contrastive loss $\mathcal{L}_{cl}$ and segmentation loss $\mathcal{L}_{seg}$:
\begin{equation}\label{eq:12}
\mathcal{L} =  \mathcal{L}_{cl} + \mathcal{L}_{seg}.
\end{equation}

During the inference, we predict the mask based on the referent token's segmentation probabilities of the last decoder layer, \ie, $Z_4^{(1)}$.

\section{Experiments}
\label{sec:experiment}
\subsection{Datasets and Implementation Details}
\label{sec:dataset}

\noindent \textbf{Datasets.} We conduct experiments on four common benchmark datasets, RefCOCO~\cite{yu2016modeling}, RefCOCO+~\cite{yu2016modeling}, G-Ref~\cite{nagaraja2016modeling,mao2016generation}, and ReferIt~\cite{kazemzadeh2014referitgame}. 
The images of the first three datasets are all based on MSCOCO~\cite{lin2014microsoft}, but are annotated with different settings. RefCOCO has a short average description length of 3.5 words, RefCOCO+ is limited to not describing absolute locations of referents, and G-Ref has a longer word count per expression (8.4 words). We follow previous works~\cite{liu2017recurrent, yang2021bottom} to split the RefCOCO and RefCOCO+ into training, validation, testA and testB. For the G-Ref, we apply both partitions of UMD and Google for the evaluation. \textcolor{black}{In addition, the ReferIt dataset, which is also the main benchmark for referring image segmentation, includes 19,894 images sourced from the IAPR TC-12~\cite{escalante2010segmented}.}

\noindent \textbf{Implementation Details.} Following~\cite{yang2022lavt}, our visual encoder is pre-trained on ImageNet22K~\cite{deng2009imagenet}, text encoder is initialized with the weights from HuggingFace~\cite{wolf2020transformers}, and image size is $480\times480$. The hyperparameters, $C^v_i$, $C^l$ and $C^t$ are $1024/2^{i-1}$, $768$ and $512$, respectively. We adopt AdamW~\cite{loshchilov2017decoupled} as the optimizer with initialized learning rate 1$e$-4 and train the model for $50$ epochs with batch size $64$. All experiments are conducted on NVIDIA Tesla A40 GPUs. Following~\cite{yang2022lavt}, we adopt \textcolor{black}{overall IoU (oIoU), mean IoU (mIoU)}, and precision at the $0.5$, $0.7$, and $0.9$ thresholds of IoU as our main evaluation metrics.

\noindent \textbf{Implementation for Generalization.} We introduce new splits \textcolor{black}{on RefCOCO series datasets} to validate the generalization, inspired by~\cite{sadhu2019zero}. Specifically, we split them according to the splits of \textcolor{black}{seen and unseen} classes on MSCOCO of open-vocabulary detection~\cite{zareian2021open}. Image-text pairs in the original training sets whose referent categories belong to the \textcolor{black}{seen} classes are selected as the new training sets. 
Likewise, the image-text pairs of test sets are also split into seen and unseen subsets according to whether the categories of the referents belong to \textcolor{black}{seen} classes. And the categories of referents in the unseen splits are not seen in the training stage. 
We consider that CRIS~\cite{wang2022cris} employs CLIP~\cite{radford2021learning} as the encoder network for transferring the knowledge of CLIP to achieve text-to-pixel alignment. Therefore, for the generalization experiments, we also take the text encoder of CLIP as the language encoder for our CGFormer and LAVT~\cite{yang2022lavt} for a fair comparison. For all methods, we train $50$ epochs using the official code and select best-performing models on the validation set for comparison. We use mIoU as the evaluation metric to eliminate the influence of categories because referents with different categories differ in size.

\subsection{Comparison with State-of-the-Art Methods}
As shown in Table \ref{tab:1} and Table \ref{tab:gen}, we compare CGFormer with state-of-the-art methods\cite{yang2022lavt,wang2022cris,kim2022restr,ding2021vision} on the four benchmarks and validate its generalization ability on our split datasets. CGFormer outperforms state-of-the-art methods on all the splits on the three datasets consistently.

\begin{table}[t]
   \centering
   \small
   \setlength{\tabcolsep}{1.9mm}{
   \begin{tabular}{l|l|c|c}
      \toprule[1pt]
      \multirow{2}{*}{\makebox[0.09\textwidth][l]{Dataset}} &\multirow{2}{*}{\makebox[0.065\textwidth][l]{Method}} &
      \multicolumn{1}{c|}{val}  & \multicolumn{1}{c}{test} \\
                                    & &\textit{ } \textit{seen} \textit{ } \textit{unseen} &\textit{ } \textit{seen} \textit{ }  \textit{unseen}  \\
      \hline
      &CRIS~\cite{wang2022cris}  &  68.66  \textit{ } 52.77 & 52.77 \textit{ } 52.66 \\
      RefCOCO&LAVT~\cite{yang2022lavt} & 73.05 \textit{ }  61.35 & 72.31 \textit{ } 57.66 \\
      &Ours &\textbf{75.52} \textit{ } \textbf{63.17} & \textbf{74.63} \textit{ } \textbf{59.03} \\
      \hline
      &CRIS~\cite{wang2022cris}   & 61.49 \textit{ } 48.08 & 60.46 \textit{ } 45.26 \\
      RefCOCO+&LAVT~\cite{yang2022lavt}  & 61.17 \textit{ } 41.49 & 60.97 \textit{ } 38.67 \\
      &Ours & \textbf{67.44} \textit{ } \textbf{51.24} & \textbf{66.35} \textit{ } \textbf{48.11} \\
      \hline
      &CRIS~\cite{wang2022cris}  & 58.64 \textit{ } 42.63 & 59.68 \textit{ } 38.88 \\
      G-Ref(U)&LAVT~\cite{yang2022lavt} & 60.16 \textit{ } 42.33 & 60.37 \textit{ } 41.38\\
      &Ours &\textbf{65.60} \textit{ } \textbf{46.11} & \textbf{65.67} \textit{ } \textbf{42.31} \\
      \hline
      &CRIS~\cite{wang2022cris}  &  42.36 \textit{ } 32.84  & \multicolumn{1}{c}{\multirow{3}{*}{$\backslash$}} \\ 
      G-Ref(G)&LAVT~\cite{yang2022lavt} & 57.33 \textit{ } 40.43 &   \\
      &Ours & \textbf{62.85} \textit{ } \textbf{45.05} &    \\
      \bottomrule[1pt]
   \end{tabular}}
   \caption{Comparison for generalization setting on the validation and test sets of RefCOCO, RefCOCO+ and G-Ref datasets using mean IoU(\%). (U): UMD partition. (G): Google partition. }
   \label{tab:gen}
\end{table}

\noindent\textbf{Comparison on Referring Image Segmentation.}
Table~\ref{tab:1} illustrates the comparison on common splits. Our CGFormer improves the average oIoU by $1.78\%$, $2.35\%$, $2.82\%$ and $6.66\%$ on RefCOCO, RefCOCO+, G-Ref, and ReferIt datasets respectively, compared to the previous best-performing methods~\cite{yang2022lavt, feng2021encoder}. This demonstrates that our object-aware reasoning and joint decoding not only achieve a better understanding of the location and appearance information in RefCOCO but also adapt to the various forms of expressions in RefCOCO+ and G-Ref.

Besides, the following three comparisons show the effectiveness of our CGFormer from different perspectives: (1) CRIS~\cite{wang2022cris} is the recently proposed CLIP-based pixel-level contrastive learning method. Compared to CRIS, CGFormer achieves clear performance improvements of $6.40\%$, $6.67\%$ and $7.59\%$ on the three RefCOCO series datasets, which indicates that our mask-level contrastive framework is more capable than the pixel-level alignment. 
(2) Compared to other methods that capture object-level information, such as MAttNet~\cite{yu2018mattnet} and BUSNet~\cite{yang2021bottom}, CGFormer significantly surpasses them by $13.96\%$ and $15.58\%$ in terms of average oIoU on RefCOCO and RefCOCO+ datasets, respectively. These results imply that our end-to-end token-based object information capturing is more simple and effective. 
(3) Moreover, we improve the performance of VLT~\cite{ding2021vision} by $10.76\%$, $13.33\%$, and $12.88\%$ on the three RefCOCO series datasets, respectively. VLT also adopts query tokens to model the object-level information, however, it does not consider the inherent differences between tokens. The large gains show the superiority of our grouping strategy cooperated with contrastive learning to distinguish tokens.
\begin{table}[]
\centering
\small
\renewcommand{\arraystretch}{1.07}
\tabcolsep=0.143cm
\begin{tabular}{l|l|ccc|c}
\toprule[1pt]
\multicolumn{1}{c|}{} &Method &\multicolumn{1}{c}{P@0.5} &\multicolumn{1}{c}{P@0.7}& \multicolumn{1}{c|}{P@0.9} &\multicolumn{1}{c}{oIoU} \\
\hline
1&baseline                                  & 75.31    & 61.48    & 16.85       & 65.70       \\ 
2&1+one token                                & 77.28    & 64.94    & 19.47      & 66.39       \\
3&1+$N$ tokens                             &77.70    & 65.12    & 19.44       &66.46 \\ 
4&3+grouping                               & 83.94    & 72.09    & 23.43       & 70.81       \\
5&4+hard assignment                    & 84.59    & 74.92    & 33.75      & 72.44       \\ 
\hline
6&5+multi-scale                          & 85.80    & 76.31    & 35.35      & 73.28        \\ 
7&5+CD (ours full)                                       & \textbf{87.23}    & \textbf{78.69}    & \textbf{38.77}      & \textbf{74.75} \\ 
\hline
8&VLT(Swin-B+BERT)$^*$                                & 83.24    & 72.81    & 24.64     & 70.89      \\
\hline
9&w/o $\mathrm{cos}$                                 & 85.64    & 76.23    & 33.96      & 73.37       \\
10&w/o learnable $\tau$                     & 86.14    & 76.99    & 36.48       & 73.50       \\ 
\bottomrule[1pt]
\end{tabular}
\caption{Ablation study on the validation set of RefCOCO. CD: Consecutive Decoder. $\mathrm{cos}$: cosine similarity operation. $\tau$: learnable parameter in Gumble Softmax. Results with $^{*}$ refer to \cite{yang2022lavt}.}
\label{tab:abl}
\vspace{-0.1cm}
\end{table}

\begin{figure*}[th]
\centering
\includegraphics[width=0.99\textwidth]{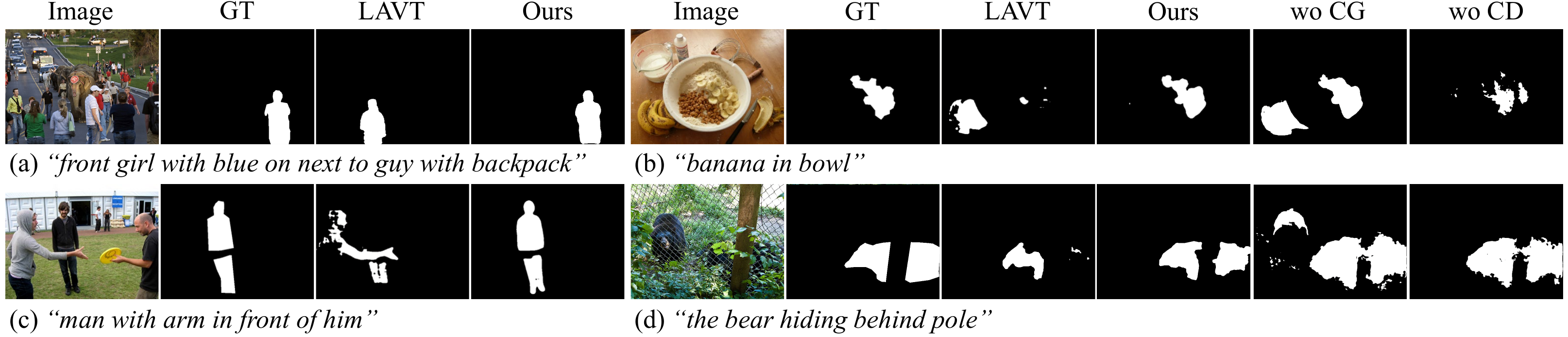}
 \vspace{-0.1cm}
\caption{\textcolor{black}{Visualization results of our CGFormer, its variants, and LAVT~\cite{yang2022lavt}.} CG: Contrastive Grouping. CD: Consecutive Decoder.}
\label{fig:vz}
 \vspace{-0.1cm}
\end{figure*}

\begin{figure}[t]
\centering
\includegraphics[width=0.99\columnwidth]{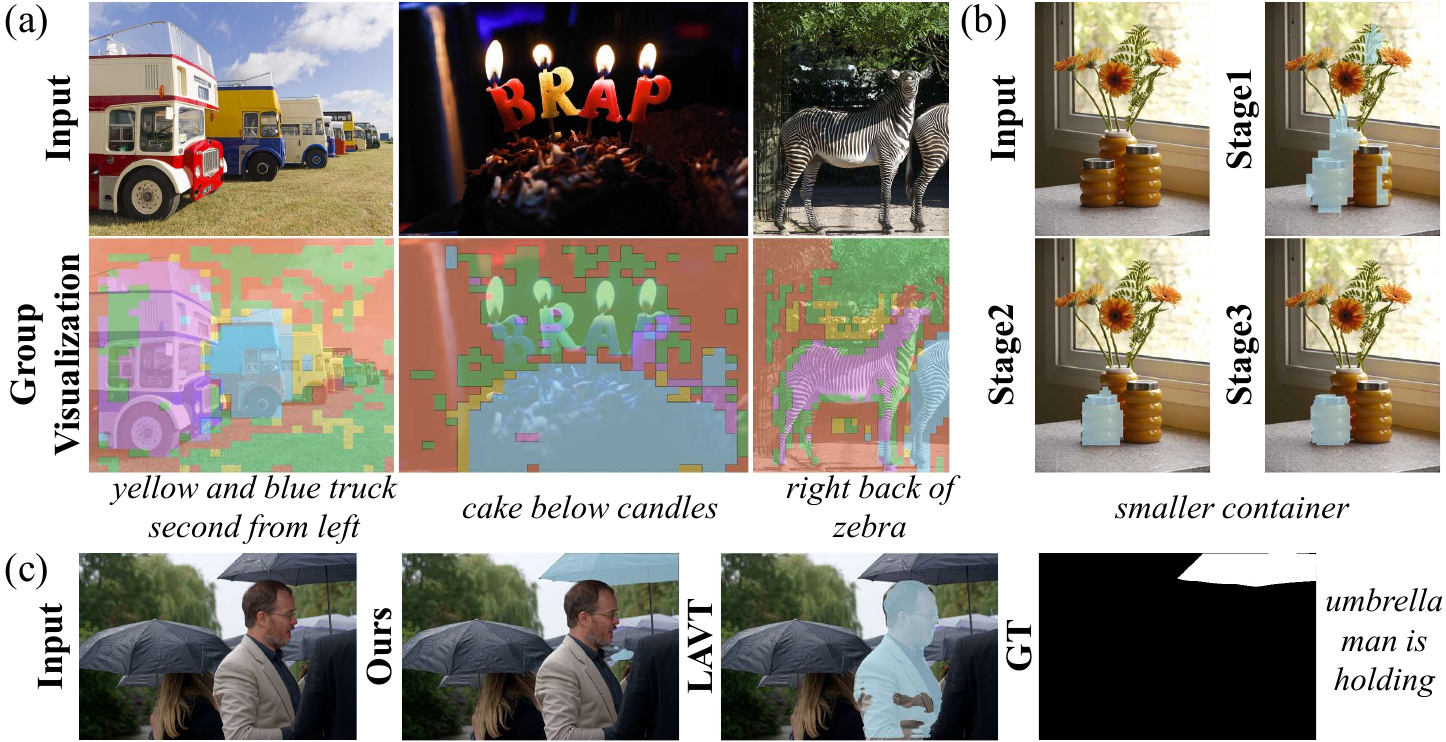}
\caption{Visualization of grouping results for (a) different tokens (in different colors), (b) the referent token in three stages and (c) segmentation results of unseen objects.} 
\label{fig:group}
\vspace{-0.2cm}
\end{figure}

\noindent\textbf{Comparison on Generalization.} 
We compare CGFormer with LAVT~\cite{yang2022lavt} and CRIS~\cite{wang2022cris} as LAVT is the current best-performing model in referring image segmentation and CRIS transfers the knowledge of the strong generalizable CLIP model~\cite{radford2021learning}. 
As is shown in Table \ref{tab:gen}, CGFormer outperforms LAVT and CRIS for both seen and unseen splits on all three datasets consistently. On the RefCOCO+ dataset that relies on understanding object-level attributes rather than location information, our performance exceeds LAVT by $5.8\%$ and $9.6\%$ in terms of average mIoU on seen and unseen splits, respectively. In addition, our CGFormer performs significantly better than other methods on the G-ref dataset with more complex languages. The mIoU of our CGFormer outperforms CIRS by $20.49\%$ and $12.21\%$ on seen and unseen splits of G-Ref(G), respectively.

\subsection{Ablation Study}
The results of the ablation study are shown in Table~\ref{tab:abl}.

\noindent\textbf{Baseline and Grouping Strategy.} (1) The baseline extracts the visual feature map $V_4$ of the visual encoder and predicts the segmentation result from the map via dynamic convolutions with the learned kernels from linguistic feature $\boldsymbol{e}$.
(2) We improve the baseline by generating a linguistic-conditioned query token and querying relevant visual features on $V_4$ to capture object-level information, which slightly improves the baseline by $0.69\%$. (3) We further extend one token to multiple tokens, but the two models have similar performance. The results suggest that simply adding tokens cannot boost performance, as these tokens are likely to focus on similar information rather than distinct regions. 
(4) Our grouping strategy cooperated with contrastive loss to make tokens can focus on different regions and let them distinguishable, which delivers a $4.35\%$ improvement. 
(5) We further use the hard assignment with learnable Gumbel Softmax to obtain a more refined grouping that achieves an improvement of $1.63\%$.

\noindent \textbf{Multi-Scale Decoding.}
We extend the single-scale model (row 5) to multi-scale one (row 6) by first parallel updating tokens at multiple levels and then integrating these tokens to predict the segmentation mask over the multi-scale visual feature map fused by FPNs~\cite{lin2017feature}. The $0.84\%$ improvement of oIoU shows the effectiveness of multi-scale features. 

We further connect grouping layers by consecutive decoding, which is applied in our final model (row 7). 
The $1.47\%$ improved oIoU of the model using a consecutive decoder (row 7) over the model using parallel querying (row 6) suggests that our joint querying and decoding in the decoder is a more desirable solution than aggregating different levels of information in parallel.

\noindent \textbf{Others.} (1) We further validate the necessity of the proposed contrastive grouping by comparing our CGFormer (row 7) with VLT~\cite{ding2021vision} (row 8) using the same visual backbone and text encoder. 
We significantly outperforms VLT by $3.99\%$, $5.88\%$, $14.13\%$ and $3.86\%$ in terms of P@0.5, P@0.7, P@0.9 and oIoU, respectively. 
(2) We replace the cosine similarity with dot produce (row 9) or fix the parameter $\tau$ in the Gumble softmax to $0.1$ (row 10), which results in a reduction of about $1.3\%$ in oIoU.

\subsection{Visualization}
Figure~\ref{fig:vz} visualizes segmentation results. The expression in (a) refers to a girl in a complex scenario with a crowd of several dozen people. 
For (b), CGFormer accurately recognizes the challenging visual concept \textit{``sliced banana”} and distinguishes it from a similar object with the same visual concept. The (c) demonstrates that our joint grouping and decoding comprehensively understand the expression and image rather than only focusing on local information \textit{``hands"} and \textit{``man"}. The (d) illustrates that CGFormer entirely segments the \textit{``bear"} even though it is shaded by \textit{``pole"} thanks to our object-aware reasoning. 

\noindent\textbf{Variant Results} of our models without the Contrastive Grouping and Consecutive Decoder are shown in the two rightmost columns of Figure~\ref{fig:vz}, respectively. Contrastive Grouping captures object-level information to distinguish between similar objects, and the Consecutive Decoder help obtain more precise segmentation results. 

\noindent\textcolor{black}{\textbf{Qualitative results of grouping} are shown in Figure~\ref{fig:group} (a) and (b), which demonstrates: (1) Both the referent token and others represent certain meaningful objects/regions. (2) Tokens can partition the objects of the same categories (e.g., the different trucks) and different categories (e.g., the cake and candles). (3) The grouping can be more precise in multiple stages (see results for referent token in b).}

\section{Conclusion}
This paper proposes a novel Contrastive Grouping with Transformer network (CGFormer) for referring image segmentation, which achieves object-aware cross-modal and cross-level reasoning. The experimental results demonstrate the superiority and the generalization ability of the proposed CGFormer compared to state-of-the-art methods.

\noindent\textbf{Acknowledgment:} this work was supported by the National Natural Science Foundation of China (No.62206174), the Shanghai Pujiang Program (No.21PJ1410900), and the Shanghai Frontiers Science Center of Human-centered Artificial Intelligence (ShangHAI).

{\small
\bibliographystyle{ieee_fullname}
\bibliography{egbib}
}

\end{document}